\documentclass[11pt]{article}
\PassOptionsToPackage{table}{xcolor}
\usepackage{acl}
\usepackage{times}
\usepackage{ulem}
\usepackage{graphicx} 
\usepackage{multirow}
\usepackage{booktabs}
\title{Evaluating VQA in Vision Language Models using Cooperative Principles}
\author{Monika Shah, Sudarshan Balaji, Somdeb Sarkhel, Sanorita Dey, Deepak Venugopal \\
University of Memphis, TN, USA \\
Adobe Research, San Jose, CA, USA \\
University of Maryland Baltimore County, USA \\ \{mshah2, sbalaji, dvngopal\}@memphis.edu, sarkhel@adobe.com, sanorita@umbc.edu}

\date{}
\newcommand{\eat}[1]{}

\begin{document}

\maketitle
\begin{abstract}
    We evaluate the performance of Vision Language Models in Visual Question Answering (VQA) when questions violate Grice's maxims. To do this, we use VLMs to generate question modifiers that add non-essential, ambiguous or false information and show that in the presence of such violations, the VLMs that we evaluate (ChatGPT, Claude, Gemini and Llava) show diminished performance. Further, we empirically show the difference between how humans reason pragmatically compared to VLMs, and the difference in VLM reasoning when it resolves violations that are human-induced compared to those that are AI-generated. Finally, we show that human cognitive effort (measured through time-on-task in an experiment) is lower for resolving VLM-induced violations, but VLMs themselves perform less accurately in such cases.
\end{abstract}
\section{Introduction}

Vision Language Models (VLMs)~\cite{team2024gemini,liu2023visual,hurst2024gpt} 
that unify Large Language Models with computer vision have made significant advances in multimodal tasks such as image captioning~\cite{yang2019auto, cornia2020meshed, wang2022end} and visual question answering (VQA)~\cite{antol2015vqa}. 
While recent advances have demonstrated the impressive potential of these models, their reasoning capabilities—and, more importantly, their limitations—are still only partially understood~\cite{campbell2024understanding}. 
In this work, we empirically analyze the sensitivity of VLM responses in VQA when there are possible violations of the \textit{Cooperative principle}~\cite{grice1975logic} that governs human pragmatic reasoning.

Specifically, Grice’s maxims explain the Cooperative principle in pragmatics. Compliance with these maxims ensures that human conversation is concise, clear, truthful and relevant, thereby ensuring that the conversing parties arrive at a mutual understanding of the discussion. In that regard, as humans begin to work with VLMs in a collaborative manner, it is important that they understand what humans say (and mean) accurately.
Prior studies have shown that, even when Grice's maxims are violated, humans are generally able to comprehend the conversation~\cite{davies2000grice}. Nevertheless, such violations impose a greater cognitive burden on the listener, as additional effort is required to infer the intended meaning~\cite{jacquet2018gricean}. In this study, we evaluate if VLMs have similar \textit{pragmatic reasoning} abilities needed to resolve such violations.


We investigate the extent to which violations of Grice’s maxims influence the reasoning processes of VLMs. Specifically, we probe VLMs with VQA questions where \textit{question modifiers} are added to human-crafted VQA questions.
While an AI model could potentially benefit from the additional information provided by modifiers, processing them will require increased pragmatic reasoning from the model. Specifically, given the multimodal nature of VQA, to resolve violations, VLMs can use context from one modality (image) to resolve violations that affect the other (text).
We generate the modified questions themselves using VLMs and use \textit{human-as-a-judge} to supervise the modification. 
Using this, we try to understand pragmatic reasoning in VLMs by answering the following questions, 
\begin{itemize}
    \item \textit{How do modifiers affect the accuracy of VLM responses in VQA?}
    \item \textit{Is there a difference in VLM reasoning when violations to Grice's maxims in VQA occur more naturally, i.e., when humans violate these maxims as opposed to AI-generated violations?}
    \item \textit{What is the relationship between human cognitive effort/reasoning and VLM pragmatic reasoning?}
\end{itemize}




We perform experiments on ChatGPT4o, Claude Sonnet 3.5, Llava 7B and Gemini-flash-1.5 using the questions from the VQA2.0 dataset~\cite{goyal2017making} as the base questions for our modifications. Our results show that the accuracy of VLMs decrease in the presence of violations, VLMs perform better on more natural violations (human-like) than artificially generated ones, and there is a significant difference in how humans and VLMs resolve violations. 
\section{Related Work}


Several datasets have been developed in VQA evaluation~\cite{antol2015vqa,goyal2017making,selvaraju2020squinting,tan2019lxmert} that aim to test the accuracy of models performing this task. The use of LLMs-as-a-judge in VQA has been growing recently~\cite{zhou2023lima},\cite{manas2024improving}.
To more deeply understand the reasoning process in VQA models, \cite{wang2024haloquest} developed a hallucination dataset consisting of questions with false premise, and visually challenging and insufficient context. 
\cite{wu2024autohallusion} developed a hallucination dataset that consists of synthesized image and question pairs. 
\cite{ribeiro2018semantically} added semantically equivalent adversaries in the questions to induce changes in the model's prediction. The work in ~\cite{britton2022question} is related to our approach where it perturbs/modifies VQA questions and analyzes its effect on LXMERT~\cite{tan2019lxmert}. In this work, we build on this modification mechanism to evaluate VLM reasoning more broadly grounded in cooperative principles.




\section{Evaluation Procedure}

 
\label{sec:evalprocedure}
Grice’s maxims characterize the fundamental principles of effective and cooperative communication. The maxim of quality suggests truthfulness, quantity suggests conciseness, relation emphasizes relevance and manner emphasizes clarity. Utterances by a speaker in a conversation are said to be {\it Grice-optimal} if they adhere to Grice's maxims.
However, in everyday conversations, Grice’s maxims are frequently violated, yet humans are typically able to reason pragmatically and infer the intended meaning~\cite{davies2000grice}. In cognitive neuroscience, it is well-established that reasoning pragmatically is associated with executive functions such as {\it selective attention}~\cite{}. Specifically, filtering information rather than processing a greater amount of information is essential to perform higher-level reasoning~\cite{}. To diagnose human learning/misconceptions, a technique widely used in educational research is to use {\it distractors}. Specifically, introducing subtle distractions (e.g. spurious information, closely-related multiple choice items, etc.) allows us to understand the depth of student understanding. Distractors create a competition for attention~\cite{}, i.e., they provide a diagnostic mechanism for understanding if the reasoning process of the learner is flexible enough to selectively attend only to cues that are relevant to achieve a goal while filtering out irrelevant cues. Here, we develop a similar diagnostic framework for VQA where we try to distract the VLM with question modifiers, thus forcing VLMs to reason pragmatically.






Grice's maxims were originally formulated to guide effective human communication in conversations. In the context of VQA, we interpret the maxims as follows. An instance ($q$,$I$), where $q$ is the question and $I$ is the associated image is said to be Grice-optimal if $q$ can be answered from $I$ without requiring additional information. Suppose we modify $q$ to $q'$ by adding information from $I$ such that the answers to ($q$,$I$) and ($q'$,$I$) are equivalent, by our definition, ($q'$,$I$) is not Grice-optimal. We add modifiers to describe two types of modifications that are significant to perception~\cite{}. Visual properties are  attributes such as color, shape, and texture of an object that can be perceived in $I$. Relational properties are spatial relationships between objects or surroundings perceived in $I$ (e.g., described with prepositional phrases such as \textit{next to}, \textit{on top of}, etc.). To diagnose the effect of modifiers, we use an approach similar to cross-validation in Machine Learning. Specifically, we add modifiers by prompting a {\it modification} VLM and evaluate its effect on a {\it response} model. 

We prompt the VLM with few-shot examples and the following rules (full prompt in appendix), i) the modified question needs to have the same answer as the original question, ii) the question type should not be altered (e.g. a \textit{what} question needs to remain a \textit{what} question), iii) the context of the question should not be altered significantly. For example,  \textit{Is there a car?}, should not change completely to  \textit{Is there a river?}, iv) either visual of relational modifier should be added and the model should specify which one it added to reflect on its choice, and v) the modifier should not give away the answer inadvertently. 

To ensure that modifiers are semantically meaningful, we use a human to judge the modified questions generated by a modification VLM. To do this, we design an Amazon Mechanical Turk (AMT) task where we show $(q,I)$, and highlight the modification (from a specific VLM) in $q'$. We ask users if the modification in $q'$ preserves the answer in $q$ by showing the answer provided to $(q,I)$. To improve quality of the responses, we add a conditional rule where if the user selects ``yes'', they are then shown a question where they are motivated to reflect on their choice by providing the main object that they had to observe in the image to perform the validation. If the user selects ``no'', they are then shown a \textit{cloze} task, where they need to replace the modified part in $q'$ with a more correct modification. This ensures that after correction, the modification still has the same question type as the original question. We use 3 workers per task and if a majority of them choose the option that the modification in $q'$ is correct (does not alter the answer to $q$), then we keep the modified question as is. If a majority choose the option that the question modification was incorrect, then we use the modification provided by one of the users who made this choice. The full AMT design is shown in the appendix.


We sample 995 questions from the test partition of the VQA v2.0 dataset~\cite{goyal2017making} including 438 binary-response questions where the answer is yes/no, 42 quantitative-response questions where the answer is numerical, and 493 open-ended response questions (more details on types of questions are provided in the appendix). For the binary and quantitative responses, we have a unique ground truth answer. For open-ended answers, we obtain the \textit{ground-truth} answers using a AMT task where for each question, we had 3 workers provide answers. [Modification details here]

Datasets used/generated from our experiments are available here\footnote{\url{https://anonymous.4open.science/r/Question-Modifier--8672/}}.

For our analysis, we consider 3 closed-weight VLMs, GPT-4o, Gemini-1.5-Flash, and Claude-3.5-Sonnet, and an open-source VLM, Llava-7B (we refer to them as GPT, Gemini, Claude and Llava). Table~\ref{tab:binaryquantresp} shows the overall \textit{percentage decrease} in answer accuracy for binary and quantitative responses. Since these questions have a unique ground truth, we can measure the accuracy of the responses exactly. 
As shown in our results, in general, adding modifiers decreases accuracy of responses. In some cases the accuracy reduction is more pronounced (e.g. when Gemini is the modification-model, the average reduction in accuracy is over 14$\%$). On average, Claude has the smallest accuracy reduction compared to the other models ($4.82\%$) while Llava has the largest reduction in accuracy ($13.28\%$). The variance in accuracy is largest for GPT since with its own modifications, its accuracy improves but with modifications introduced by other VLMs, the accuracy greatly reduces (e.g. Gemini modifications causes the GPT accuracy to drop by over $38\%$).
To further assess the significance of modifiers, we perform the McNemar's test~\cite{trajman2008mcnemar} where we have a dichotomous response measure, i.e., the response is either correct or not. For each response-model, we assess the change in responses before and after the violations are introduced by each of the modification-models. The results of the tests are summarized in Table~\ref{tab:mcnemarbinquant}. As shown here, clearly we can reject the null hypothesis in most cases ($p<0.05$) and can conclude that modifications significantly affect the responses of all VLMs regardless of the source of the modification, i.e., whether they were introduced by themselves or by other VLMs. The notable exception is Claude where the violations seem to not change its responses significantly for 3 of the 4 modification-models.

\begin{figure}
    \centering
    \includegraphics[width=0.85\linewidth]{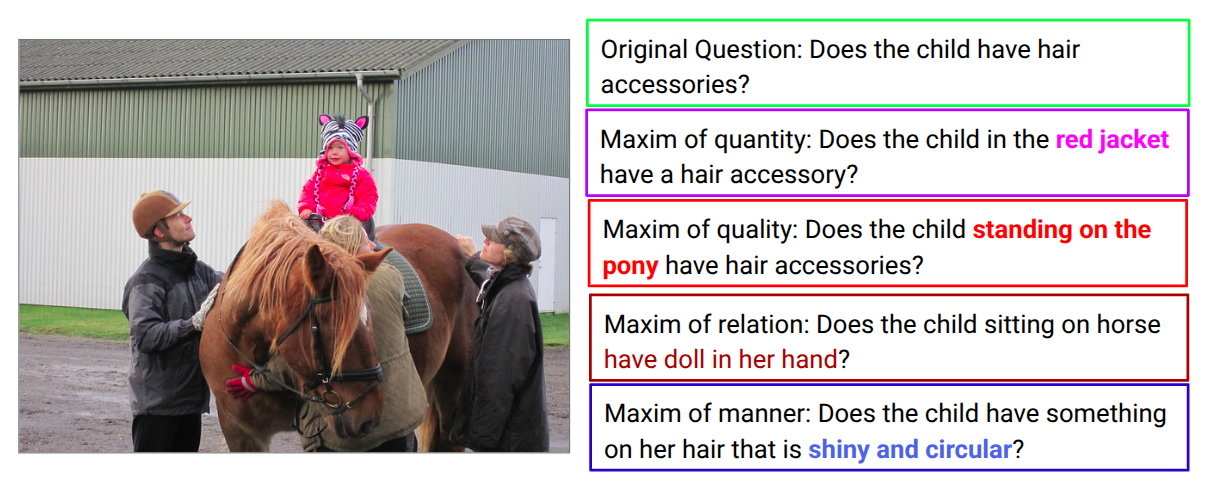}
    \caption{Illustrating violations of Grice's maxims in VQA}
    \label{fig:maxim}
\end{figure}

An example illustrating this is shown in Fig.~\ref{fig:maxim}.

\begin{figure}
    \centering
    \includegraphics[width=1\linewidth]{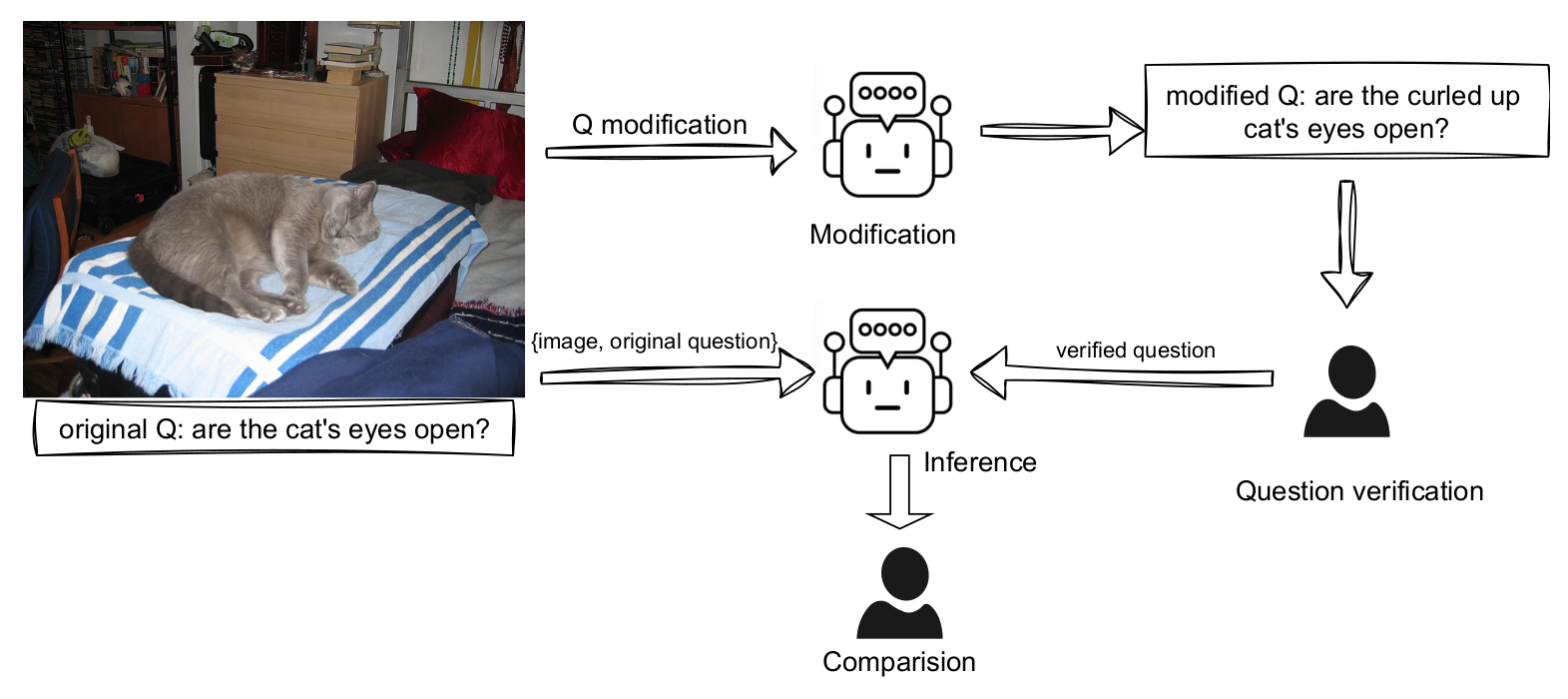}
    \caption{Illustration of the evaluation process. A VLM modifies the question which is verified by a human. A VLM answers the original and modified questions and a human compares their responses. For all the human tasks, we used Amazon Mechanical Turk (paid workers between $5-10$ cents per individual response depending on the task complexity.)}
    \label{fig:pipeline}
\end{figure}

\subsection{Adding Modifiers}

Given ($q$,$I$), where $I$ is the image and $q$ is the original human-crafted question associated with $I$, we constrain the modified question $q'$ to augment $q$ with properties related to $I$. 
We consider two properties that are broad enough to describe the image in greater detail, namely, \textit{visual properties} and \textit{relational properties}. Visual properties refer to object-level attributes—such as color, shape, and texture—that can be directly perceived in $I$. In contrast, relational properties are context-dependent, describing spatial relationships between objects (e.g., \textit{next to}, \textit{on top of}, etc.). We generate $q'$ from a VLM with few-shot prompting. The main instructions are provided below (full prompt in appendix).




\noindent{\bf Prompt.} We instruct the VLM with the following rules and few-shot examples, i) the modified question needs to have the same answer as the original question, ii) the question type should not be altered (e.g. a \textit{what} question needs to remain a \textit{what} question), iii) the context of the question should not be altered significantly. For example,  \textit{Is there a car?}, should not change completely to  \textit{Is there a river?}, iv) either visual of relational modifier should be added and the model should specify which one it added to reflect on its choice, and v) the modifier should not give away the answer inadvertently.

\eat{
\begin{figure*}
    \centering
    \includegraphics[width=0.75\linewidth]{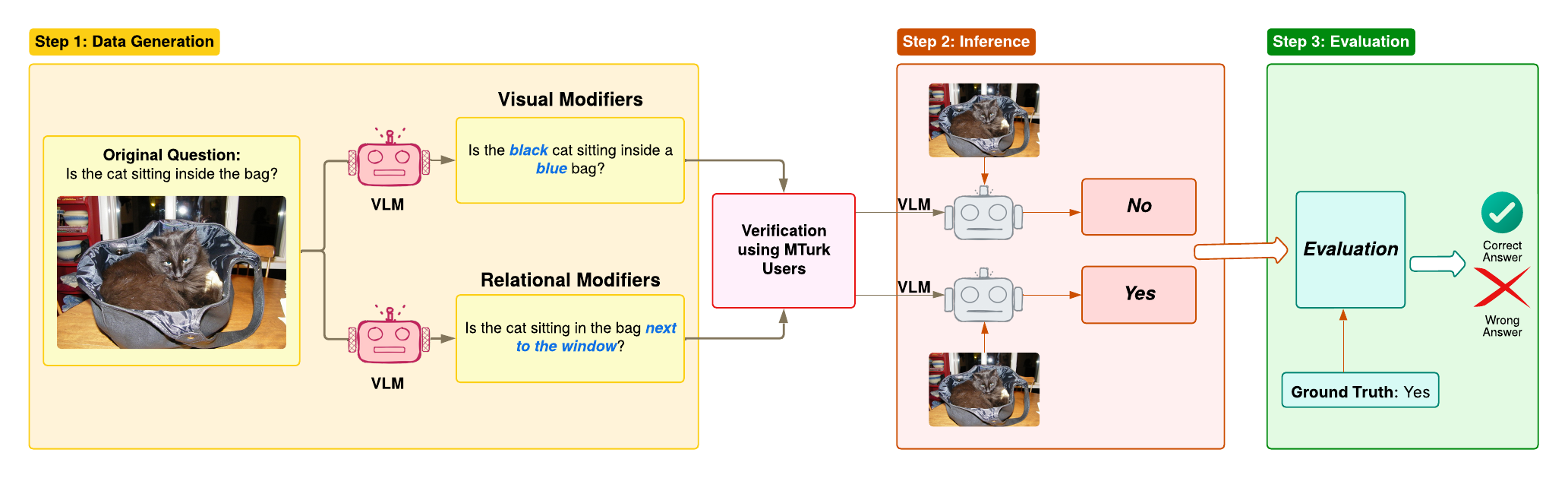}
    \caption{Caption}
    \label{fig:placeholder}
\end{figure*}
}

\noindent{\bf Human-as-a-judge.} Given an instance $(q,I)$ and $q'$ that has been modified by a specific VLM, we want to know if the violation induced by $q'$ can be pragmatically inferred and resolved by humans. To do this, we design an Amazon Mechanical Turk (AMT) task where we show $(q,I)$, and highlight the modification (from a specific VLM) in $q'$. We ask users if the modification in $q'$ preserves the answer in $q$ by showing the answer provided to $(q,I)$. To improve quality of the responses, we add a conditional rule where if the user selects ``yes'', they are then shown a question where they are motivated to reflect on their choice by providing the main object that they had to observe in the image to perform the validation. If the user selects ``no'', they are then shown a \textit{cloze} task, where they need to replace the modified part in $q'$ with a more correct modification. This ensures that after correction, the modification still has the same question type as the original question. We use 3 workers per task and if a majority of them choose the option that the modification in $q'$ is correct (does not alter the answer to $q$), then we keep the modified question as is. If a majority choose the option that the question modification was incorrect, then we use the modification provided by one of the users who made this choice. The full AMT design is shown in the appendix.

\noindent{\bf Evaluation of open-ended responses.} We develop an AMT task to evaluate open-ended responses from VLMs. Specifically, since evaluating answers independently is quite hard, we want users to perform a comparison which yields significantly better responses from AMT workers. We show the user two question answer pairs, the first one is the original question and the VLM's answer,and the second is the modified question with the VLM's answer. We ask users to compare the pairs and choose if one answer was better than the other, or both answers were equally acceptable (AMT task shown in appendix).  An illustration of our evaluation process is shown in Fig.~\ref{fig:pipeline}.

\section{VLM Performance under Gricean Violations}

\label{sec:testaccuracy}
Similar to cross-validation, we generate modified questions from 4 different VLMs, GPT-4o, Gemini-1.5-Flash, Claude-3.5-Sonnet and Llava-7B (we refer to them as GPT, Gemini, Claude and Llava), and test the accuracy of answers given by each of the VLMs. We term the VLM that performed modification as the \textit{modification-model} and the VLM that provided the answer as the \textit{response-model}. 
\begin{table*}
  \centering
  \resizebox{2.0\columnwidth}{!}{%
  \begin{tabular}{|l|l|l|l|l|l|l|l|l|l|l|}
    \hline
    {\bf \small Response-Model / Modification-Model} &
      \textit{GPT} &
      \textit{Gemini} &
      \textit{Claude} &
      \textit{LLava} & {$Average \pm std$}\\
      
    \hline
    \textit{GPT} & -3.13\% & {38.08\%} & 2.97\% & 12.10\% &$12.51 \pm 18.16\%$\\
    \hline
    \textit{Gemini} & 10.05\% & 8.60\% & 9.07\% &   12.50\% & $10.06 \pm 1.74\%$\\
    \hline
    \textit{Claude} & 2.65\% & 2.72\% & 2.43\% & 11.48\% & $4.82 \pm 4.44\%$\\
    \hline
    \textit{Llava} & 15.56\% & 9.50\% & 10.51\% & 17.56\% & \cellcolor{yellow!50}{$13.28 \pm 3.89\%$}\\
    \hline
    $Average \pm std$& $6.28\pm 8.21$\% & \cellcolor{yellow!50}{$14.73 \pm 15.86\%$} & $6.25 \pm 4.14\%$ & $13.41 \pm 2.80\%$ &\\
    \hline
  \end{tabular}
  }
  \caption{$\%$ decrease in response accuracy for modified questions compared to original questions for binary and quantitative responses (smaller is better). Rows indicate the VLM providing the response and columns the VLM performing the modification.}
  \label{tab:binaryquantresp}
\end{table*}
We categorize the questions based on response-types: (i) binary-response questions, which elicit a yes/no answer; (ii) quantitative-response questions, which require a numerical value; and (iii) open-ended response questions, which require natural language answers.
We compare VLM responses to the original human-crafted questions with their responses to modified questions. For this, we use a sample of binary, quantitative and open-ended response questions from the test partition of the VQA v2.0 dataset ~\cite{goyal2017making}. We use a total 995 questions including 438 binary-response, 42 quantitative-response and 493 open-ended questions (more details on types of questions are provided in the appendix). For open-ended answers, we obtain the \textit{ground-truth} answers using a AMT task where for each question, we had 3 workers provide answers. Datasets used/generated from our experiments are available here\footnote{\url{https://anonymous.4open.science/r/Question-Modifier--8672/}}.

\subsection{Evaluating Response Accuracy}
Table~\ref{tab:binaryquantresp} shows the \textit{percentage decrease} in answer accuracy for each of the VLMs for binary and quantitative responses. Note that answers to these questions have a unique ground truth, and therefore, we can measure the accuracy of the responses exactly. 
Our results show that overall, almost all combinations of modification-models and response-models show a percentage decrease in accuracy suggesting that modifiers negatively affect VLM reasoning in general. The notable exception was the case when GPT was both the response- and modification-model. The modification-model that induced least accurate responses was Gemini while the least accurate response-model was Llava. The difference between the column and the row averages indicate how much change a VLM induces when used to introduce violations compared to how its own responses change in the presence of violations. When ranked by this difference, we have Gemini ($4.67\%$), Claude ($1.43\%$), Llava ($0.13\%$) and GPT ($-6.22\%$)). Thus, it turns out that Gemini introduces strong violations that make VLM reasoning harder, but its own reasoning in the presence of violations is less affected. The opposite is true for GPT. Further, we see that for GPT, the standard deviation is very large for its responses ($18.16\%$), especially, the difference between the best and poorest performance is the largest for GPT since it performs best on its own modifications but very poorly on Gemini's modifications. On the other hand, Gemini yields the most consistent performance (smallest standard deviation $1.74\%$) over all modification-models, but Claude has the smallest change in accuracy ($4.82\%$).

\noindent{\bf Significance Tests.} To further assess the significance of violations, we perform the McNemar's test~\cite{trajman2008mcnemar} on the binary and quantitative responses where we have a dichotomous response measure, i.e., the response is either correct or not. For each response-model, we assess the change in responses before and after the violations are introduced by each of the modification-models. The results of the tests are summarized in Table~\ref{tab:mcnemarbinquant}. As shown here, clearly we can reject the null hypothesis in most cases ($p<0.05$) and can conclude that modifications significantly affect the responses of all VLMs regardless of the source of the modification, i.e., whether they were introduced by themselves or by other VLMs. The notable exception is Claude where the violations seem to not change its responses significantly for 3 of the 4 modification-models. In combination with the small standard deviation in accuracy (Table~\ref{tab:binaryquantresp}), this result shows that \textit{the presence of modifier-induced violations has a smaller effect on Claude responses compared to the other models}.

\begin{table*}
  \centering
  \resizebox{1.75\columnwidth}{!}{%
  \begin{tabular}{|l|l|l|l|l|l|l|l|l|l|}
    \hline
    \multirow{2}{*}{\bf \small Response-Model / Modification-Model} &
      \multicolumn{2}{c|}{GPT} &
      \multicolumn{2}{c|}{Gemini} &
      \multicolumn{2}{c|}{Claude} &
      \multicolumn{2}{c|}{LLava}\\
      
    & \multicolumn{1}{c}{\it Statistic} & \multicolumn{1}{c|}{\it P-value} & \multicolumn{1}{c}{\it Statistic} & \multicolumn{1}{c|}{\it P-value} & \multicolumn{1}{c}{\it Statistic} & \multicolumn{1}{c|}{\it P-value} & \multicolumn{1}{c}{\it Statistic} & \multicolumn{1}{c|}{\it P-value}\\
    \hline
    GPT & 4.84 & 0.028 & 15.05 & 0.00 & 3.45 & 0.063 & 9.25 & 0.002 \\
    \hline
    Gemini & 13.78 & 0.00 & 9.84 & 0.002 & 11.22 & 0.001 & 9.25 & 0.002\\
    \hline
    Claude & { 1.66} & 0.197 & {1.515} & {0.218} & 1.80 & 0.18 & {8.89} & {0.003}\\
    \hline
    Llava & { 27.50} & {0.00} & {14.45} & {0.00} & {20.57} & {0.00} & {15.29} & {0.00}\\
    \hline
  \end{tabular}
  }
  \caption{Mcnemar's test for assessing significance of the modifiers added to binary and quantitative questions.}
  \label{tab:mcnemarbinquant}
\end{table*}

\begin{table*}
  \centering
  \resizebox{1.75\columnwidth}{!}{%
  \begin{tabular}{|l|l|l|l|l|l|l|l|l|l|l|}
    \hline
    {\bf \small Response-Model / Modification-Model} & {GPT-4o} & {Gemini-1.5-Flash} & {Claude-3.5-Sonnet} & {LLava-7B} & {$Average \pm std$}\\
      
    \hline
    GPT4o & 7.68\% & 15.55\% & 12.32\% & 15.96\% & $12.88 \pm 3.32$\%\\
    \hline
    Gemini & 16.15\% & 18.46\% & 16.15\% & 20.0\% & $17.69 \pm 1.63\%$\\
    \hline
    Claude & { 21.53\%} & {20.76 \%} & 13.07\% & {16.92\%} & $18.07 \pm 3.37$\%\\
    \hline 
    Llava & {25.38\%} & {20.00\%} & {20.00\%} & {18.46\%} & \cellcolor{yellow!50}$20.96 \pm 2.63$\% \\
    \hline
    $Average \pm std$& $17.69 \pm 6.64$\% & \cellcolor{yellow!50}$18.69 \pm 1.99$\% & $15.39 \pm 3.03$\% & $17.84 \pm 1.54$\% & \\
    \hline
  \end{tabular}
  }
  \caption{Comparing open-ended responses before and after modifications. The $\%$ of questions where human users in AMT chose that the response after modification is \textit{worse} than the response before modification (for each question, we have 3 workers and we use the majority choice).}
  \label{tab:openendedhumaneval}
\end{table*}

\noindent{\bf Open-ended Responses.} For open-ended responses, we use the AMT-based human evaluation described in Section~\ref{sec:evalprocedure}. Table~\ref{tab:openendedhumaneval} summarizes our results from this study where we show the percentage of open-ended cases where the model's response to the modified question was worse than the model's response to the original question (we average this among the 3 ground truth answers). Interestingly, when we consider the maximum of row and column averages, the results are identical to the results for the binary \& quantitative responses. Specifically, Gemini as the modification-model introduces violations that make reasoning harder across all response-models. Llava generates open-ended responses that are worse in the presence of violations compared to all other VLMs. The difference between row and column averages are as follows, GPT ($4.81\%$), Gemini ($1\%$), Claude ($-2.68\%$) and Llava ($-3.12\%$). Thus, unlike the binary \& quantitative responses, here GPT introduces violations that are harder to answer (also has the largest standard deviation among all modification-model column, $6.64\%$) and its own responses to modified questions are less affected by the modifiers. Since GPT is stronger in terms of text generation, it would seem like it can produce open-ended responses better than constrained responses. As a sanity check to the human evaluation results, we perform the following experiment. We ask AMT-users to answer the open-ended modified questions (we collect 3 responses for each question). We then match the responses provided by the VLMs to these 3 responses and if there is an exact match, we consider that the VLM response is accurate. We then compute the correlation between the accuracy results of exact matches with the human evaluation results in Table~\ref{tab:openendedhumaneval}. The correlations for response-models (column-wise) are GPT ($0.377$), Gemini ($-0.99$), Claude ($0.83$), Llava ($0.87$). The row-wise correlations are GPT ($0.62$), Gemini ($0.81$), Claude ($0.71$), Llava ($-0.28$). The overall correlation is 0.21 which indicates positive correlation.
\subsection{Pragmatic Reasoning under Unreliable Information}

In the previous section, we considered cases where the modification-model adds modifiers with the goal of preserving the same answer as the original question. This means that we mainly focus on modifiers that may add ambiguous or non-essential information but do not add information that is known to be untruthful or irrelevant. Thus, they are more likely to violate the maxims of quantity and manner but are less likely to violate the maxims of quality and relation. When false or irrelevant information appears, human pragmatic reasoning often uses context and common-sense knowledge to recognize it as unreliable. In cases where the false/irrelevant information does not make sense, humans usually default back to the intended meaning by inferring it instead of trusting false information literally. To evaluate what VLMs do in this case, we add modifiers that intentionally add false or irrelevant information, and therefore are likely violate the maxims of quality and relation. Specifically, we prompt each of the modification-models with the original question without providing it an image associated with the question, and task it to add modifiers to the question. Thus, the model \textit{hallucinates} a visual or relational modifier for the question since it does not have the context of the image. To reason pragmatically, the VLM providing the response should be able to use information from one modality (the image) to correct information in the other (the text). We use the same dataset as in the previous section, and ask each VLM to add modifiers without providing it an image in the prompt.


\begin{figure}
    \centering
    \includegraphics[width=0.85\linewidth]{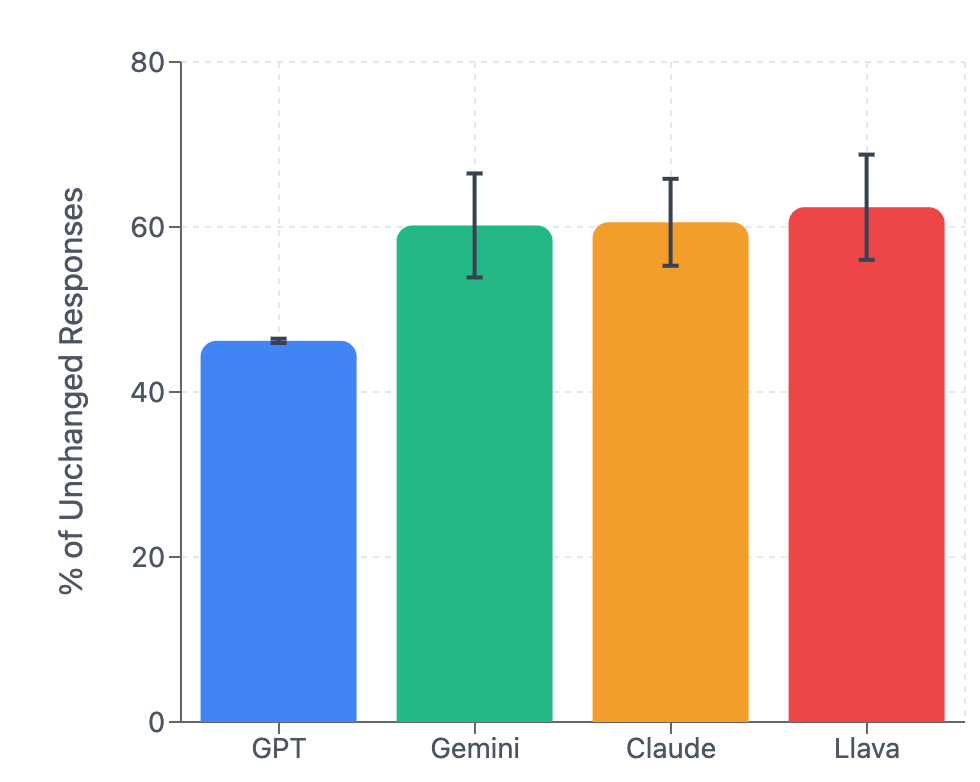}
    \caption{Comparing the $\%$ of questions where the response to a question is unchanged after a hallucinated modifier is added. Results shown for binary and quantitative response types. The y-axis shows the average (and standard deviation) over 4 hallucinated modifications each added using one of the VLMs (graph created with Claude).}
    \label{fig:randommod}
\end{figure}


Fig.~\ref{fig:randommod} compares how VLMs react to the presence of hallucinated modifiers in binary \& quantitative responses. For hallucinated modifiers added from each of the modification-models, we compute the $\%$ of questions where a VLM's response remained unchanged compared to its response for the original question. This corresponds to cases where the VLM could pragmatically reason, i.e., it could ignore the false/irrelevant information in the modifier and therefore had the same response as before the modifier was introduced. Fig.~\ref{fig:randommod} shows the average and standard deviation of results over all the modification-models. As seen here, GPT had a smaller $\%$ of instances where it had an unchanged response to hallucinated modifiers (around 43$\%$), while for all the other models it was greater than $60\%$ of questions. Thus, it could be that GPT pays attention to the text in the question more literally compared to the others. In the other cases, the models possibly used the context of the image in the question to parse out information that was relevant or false. 

\begin{table}
  \centering
  \resizebox{\columnwidth}{!}{%
  \begin{tabular}{|l|l|l|l|l|l|}
    \hline
    {\bf Response Type} &{GPT} & {Gemini} & {Claude} &{Llava}\\
    \hline
    Binary \& Quantitative & 83.53\% & 80.83\% & 83.87\% & 77.44\%\\
    \hline
    Open-ended & 0.874 & 0.864 & 0.864  & 0.865 \\
    \hline
  \end{tabular}
  }
  \caption{Comparing VLM and human pragmatic reasoning. For binary \& quantitative responses, we show the $\%$ of cases where the VLM and human answers match for hallucinated modifications. For open-ended responses, we show the BERTScore (F1-score) between human and VLM responses.}
  \label{tab:human-random-compare}
\end{table} 

\eat{
\begin{table}
  \centering
  \resizebox{\columnwidth}{!}{%
  \begin{tabular}{|l|l|l|l|l|l|}
    \hline
    {\bf BERTScore} &{GPT} & {Gemini} & {Claude} &{LLava}\\
    \hline
    Precision & 0.855 & 0.853 & 0.853 & 0.859 \\
    \hline
    Recall & 0.896 & 0.876 & 0.876 &  0.872\\
    \hline
    F1-Score & 0.874 & 0.864 & 0.864  & 0.865 \\
    \hline
  \end{tabular}
  }
  \caption{BertScore}
  \label{tab:human-random-compare}
\end{table}
}

Next, we compare how VLMs perform in relation to human pragmatic reasoning. Specifically, we ask humans to answer the same questions as before containing hallucinated modifiers. For each question, we ask 3 AMT workers to answer the question. Note that we do not tell users that there may be mistakes within the questions (e.g. false information). We want to know if the way human pragmatic reasoning uses multiple modalities where they automatically filter out noise in one modality with information from the other is similar to that of VLMs. Table~\ref{tab:human-random-compare} compares human and VLM responses to questions with hallucinated modifications. For binary \& quantitative responses, we show the $\%$ of cases where the responses of VLMs match that of humans (for each question, we choose the human response as the majority response given by the 3 human users). As seen from our results, GPT, Gemini, and Claude match with human responses better than Llava. For open-ended responses, we compute the BERTScore~\cite{Zhang*2020BERTScore:} (with default parameters) F1-score metric to measure semantic similarity between human and VLM responses (we use the average F1-score across the 3 human responses). When compared to scores computed for binary \& quantitative responses, the F1-score is higher which suggests that compared to constrained responses, in open-ended responses, VLMs are more likely to respond with answers that are semantically similar to human answers. To analyze this more deeply, we compare the binary responses of humans and VLMs in Table~\ref{tab:human-random-conditional}. We see from these results that in questions where humans gave a negative response to the binary question, VLMs seem to match their responses quite closely. However, in cases where humans gave a positive response, VLMs only give a positive response roughly in half of such questions. This indicates that in binary responses, \textit{VLM inference is more logical}, i.e., if the question is partially true, they yield a negative response. Human inference on the other hand assumes \textit{ uncertainty}, i.e., when the question is partially true, depending on the importance of that partially true content given the image context, they may respond with a positive answer (some examples in appendix).






\begin{table}
  \centering
  \resizebox{\columnwidth}{!}{%
  \begin{tabular}{|l|l|l|l|l|l|}
    \hline
    {\bf Human Response} &{GPT} & {Gemini} & {Claude} &{LLava}\\
    \hline
    Negative Response & 88.79\% & 89.47\% & 92.44\% & 86.14\%\\
    \hline
    Positive Response & 62.79\% & 48.35\% & 51.11\% & 45.05\%\\
    \hline
  \end{tabular}
  }
  \caption{Comparing binary responses of humans with VLMs. We show the \% of VLM responses aligned with the positive and negative responses of humans.}
  \label{tab:human-random-conditional}
\end{table}

\section{Comparing Human-Generated and Model-Generated Violations}

\label{sec:trainhumans}
In our experiments so far, violations to Grice's maxims are induced through modifiers generated by VLM models. However, the cooperative principles were originally formulated in the context of \textit{human-to-human} dialogue. VLMs do not inherently share the same pragmatics as humans and therefore, they are more likely to generate violations that are not \textit{human-like}. To understand the difference in VLM reasoning in the presence of violations induced by humans compared to those induced by AI, we use a dataset that consists of human-modified questions~\cite{britton2022question}. Specifically, this dataset was constructed using AMT where workers were provided instructions to add modifiers to questions from the training partition of the VQA v2.0 dataset. Note that since the training partition is used, the ground truth answers from the dataset to compute the accuracy on the modified questions. From this dataset, we sample 500 binary-response questions and 130 open-ended response questions consisting of inference questions that require deeper reasoning, namely, \textit{who, which, where, and why} questions. More details about the dataset are described in the appendix. 

\begin{figure}
    \centering
    \includegraphics[width=0.8\linewidth]{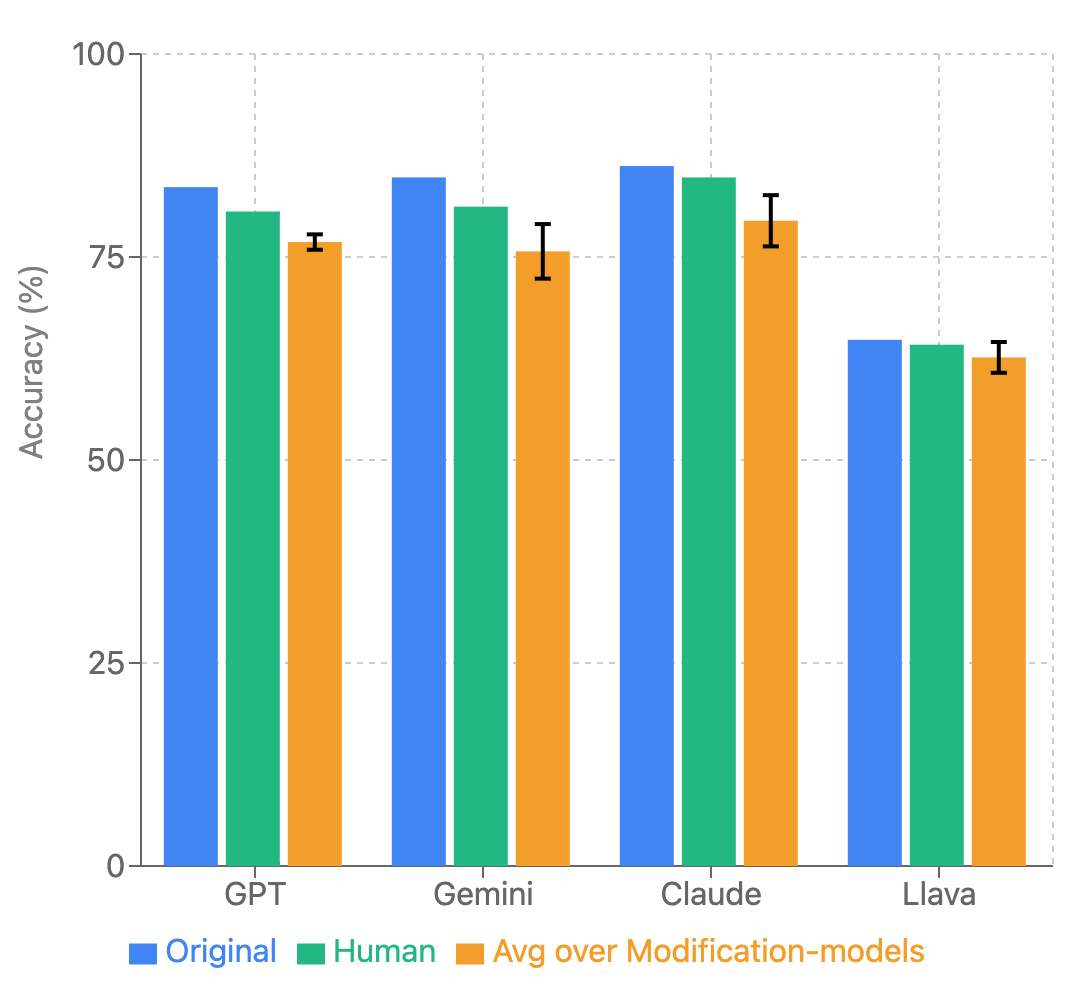}
    \caption{Comparing the accuracy of models on the original questions, questions modified by humans, and questions modified by VLMs (the average accuracy and standard deviation is shown in this case since we have modifications by 4 models). Results shown for binary responses  (graph created with Claude).}
    \label{fig:accuracy-humanAI-bin}
\end{figure}

\subsection{Comparison of Accuracy}
\begin{figure}
    \centering
    \includegraphics[width=0.8\linewidth]{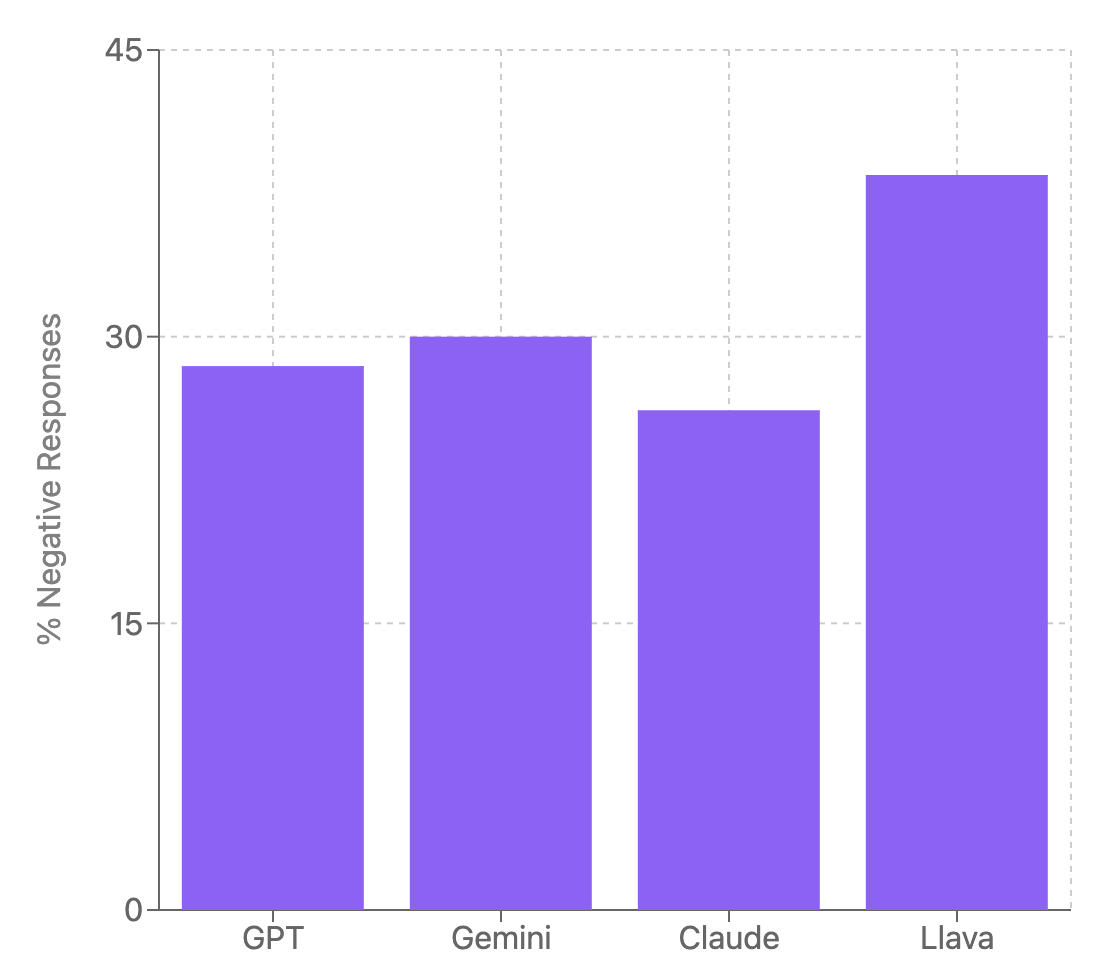}
    \caption{Comparing open-ended responses of models using human evaluation. Results show the $\%$ of cases where AMT-workers chose that the model's response to the AI-modified question was \textit{worse} than its response for the human-modified question (graph created with Claude).}
    \label{fig:accuracy-humanAI-openended}
\end{figure}

Fig.~\ref{fig:accuracy-humanAI-bin} compares the accuracies of the VLM responses (binary) on the original, human-modified and AI-modified questions. As seen here, for all 4 models, the accuracy in VLM-modified question responses is smaller compared to human-modified question responses. Claude has the best accuracy for both human and AI-modified questions, and Llava has the lowest accuracy for both. To analyze the responses for the open-ended questions, we use a AMT design similar to that in Section~\ref{sec:evalprocedure}. Specifically, we ask users to choose if they prefer the response to a human-modified question or the response to the AI-modified question. We use 3 workers for each question and use a simple majority to decide which response was preferred. Fig.~\ref{fig:accuracy-humanAI-openended} shows for each model, the $\%$ of cases where humans chose that its response to the AI-modified question was worse than its response to the human-modified question. Similar to the results in the binary/quantitative responses, the results here show that the Llava had significantly worse responses to open-ended questions when AI-models modified these questions compared to the other models. The other models had fairly similar results, and overall, Claude responses to AI-modified questions were better compared to other models ($26\%$ of responses were deemed to be worse compared to responses to human-generated modifications). 


\noindent{\bf CLIP Analysis.} To understand the difference between reasoning over AI- and human-generated violations, we analyze the alignment of modified questions with the image. Specifically, we use CLIPScore~\cite{hessel2021clipscore} (with default parameters) to measure the extent to which questions with violations are grounded in the visual content. Table~\ref{tab:clipscore} shows our results. As seen here, for both the binary response questions as well as open-ended response questions, the CLIPScore for human-modifications is different from each of the VLM-modification scores. Compared to all models, Claude has the most similar alignment ($0.639$) to human modifications ($0.612$) in binary questions. In open-ended questions, the alignment of human modifications ($0.645$) is most similar to Gemini ($0.644$). Note that Llava has a significantly different CLIPScore from the human-modified score which may explain why it also performs poorly compared to the other models. That is, it may lack the ability to ground its violations in a human-like manner and therefore its performance in the presence of violations is worse than other models. Note that the CLIPScore for Llava is the highest, however, this because it is adding significantly longer modifiers. Specifically, the average lengths of the modifications are as follows, human (2.1 words), GPT (2.5 words), Gemini (2.6 words), Claude (2.7 words) and Llava (3.8 words). Thus, Llava is adding more modifications, but its modifications are unrealistic compared to human modifications. To summarize our results, \textit{VLMs appear to handle more realistic (human-like) violations more effectively, while they struggle with artificially generated violations. Further, VLMs that are capable of generating human-like violations are also capable of reasoning more accurately in the presence of violations.}

\begin{table}
  \centering
  \resizebox{\columnwidth}{!}{%
  \begin{tabular}{|l|l|l|l|l|l|}
    \hline
    {\bf Response Type} & {Human} &{GPT} & {Gemini} & {Claude} &{LLava}\\
    \hline
    Binary \& Quantitative & 0.612 & 0.641 & 0.658 & 0.639 & 0.670\\
    \hline
    Open-ended & 0.645 & 0.657 & 0.644 & 0.654 & 0.675\\
    \hline
  \end{tabular}
  }
  \caption{CLIPScores of human and AI modifications}
  \label{tab:clipscore}
\end{table}    

\eat{
\begin{table}
  \centering
  \resizebox{\columnwidth}{!}{%
  \begin{tabular}{|l|l|l|l|l|l|}
    \hline
    {\bf Response Type} &{GPT} & {Gemini} & {Claude} &{LLava}\\
    \hline
    Yes/No & 16.47\% & 19.17\% & 16.13\% & 22.56\%\\
    \hline
    Open-ended & 0 & 0 & 0 & 0\\
    \hline
  \end{tabular}
  }
  \caption{\% change in answer when the modified questions with added random modifier are given to human vs. VLM models}
  \label{tab:clipscore}
\end{table} 
}
\subsection{VLM Reasoning in Relation to Human Cognitive Effort}


When Grice’s maxims are violated, the cognitive processing effort required for interpretation increases since context and inference must be used to recover the intended meaning. This implies that VLMs must use multimodal cues to resolve such violations and recover the intended meaning. Next, we analyze if the cognitive effort required by humans to resolve violations are indicative of VLM reasoning. Specifically, to measure cognitive load in resolving violations, we set up an experiment where we ask humans to answer a modified question. Further, we ask them to \textit{reflect} on how they answered the question by simplifying the modified question, which they demonstrate by removing parts which they do not need to arrive at a final answer (AMT task is shown in the appendix). We measure the \textit{time-on-task} to perform this activity to measure the effort it took to resolve violations in the modified questions. . 

\begin{figure}
    \centering
    \includegraphics[width=1\linewidth]{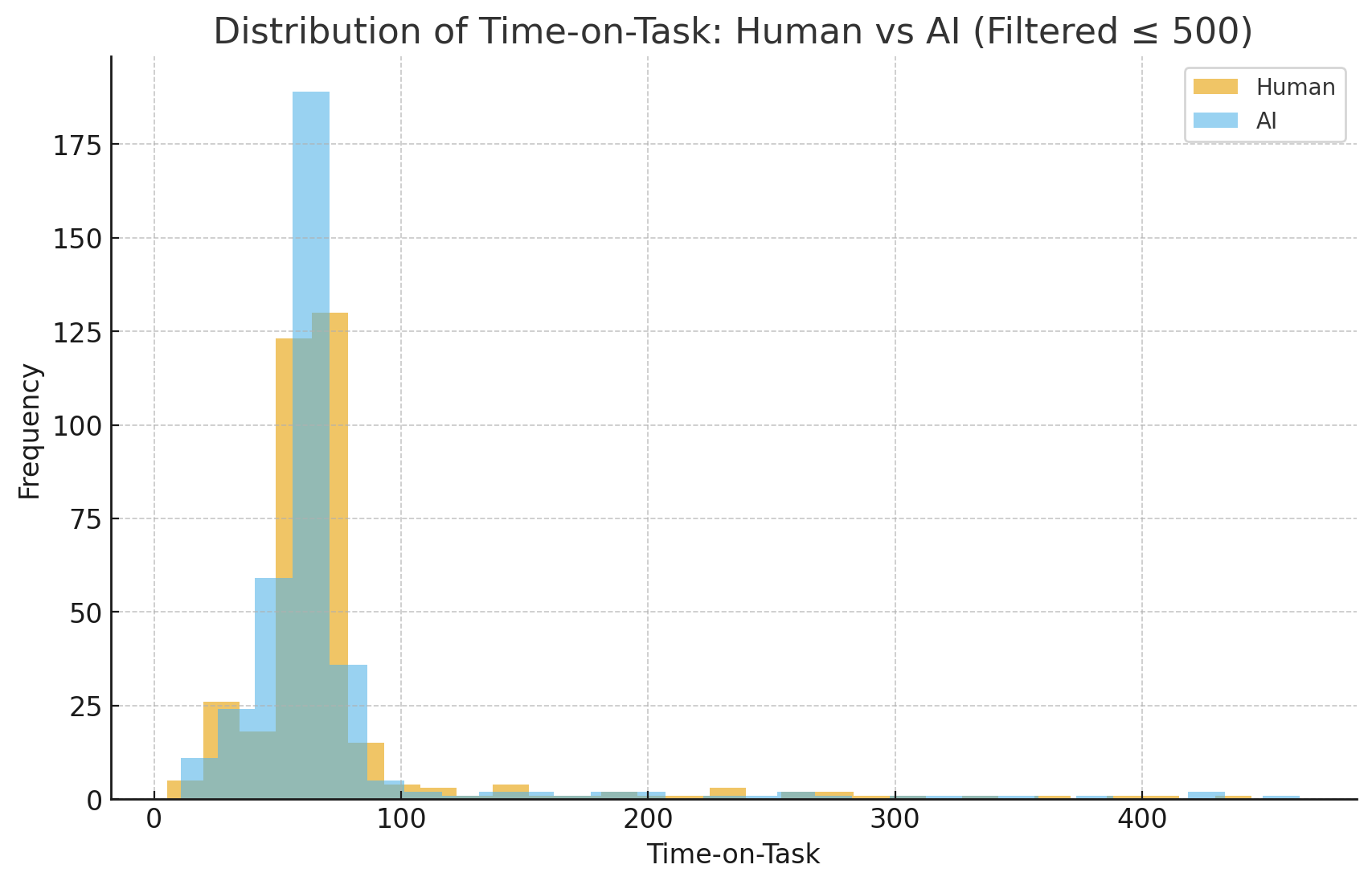}
    \caption{Distribution of time-on-task in seconds measured for the experiment where AMT workers were asked to answer modified questions and reflect on the modification by simplifying the question. Results are shown for human and AI (VLMs) modified questions (graph created with ChatGPT).}
    \label{fig:distribution-tot}
\end{figure}

We measure the time-on-task by computing the time AMT workers start the task to the time they submit their response. We filter outliers where the elapsed time is greater than 500 seconds since this could be as a result of workers starting many jobs in parallel. From the dataset used in the previous section, we sample 100 binary response questions modified by each of the VLMs and perform our analysis on a total of 400 modified questions by VLMs and the same questions modified by humans. We use 3 workers per task and use the average time-on-task for our analysis. Fig.~\ref{fig:distribution-tot} compares the distribution of the time-on-task for questions modified by humans and those modified by AI models. As shown by these results, the distribution shapes for both human and AI-modified questions are similar. The mean time taken by humans for resolving human modifications is 73.4 while for VLM modifications it is 74.2. Our results show that resolving VLM modified questions takes lower effort in humans as compared to resolving human modifications. However, as seen in the previous section, it seems like the reasoning process of VLMs answer human modifications more accurately. This seems to suggest that \textit{since VLMs are typically trained on natural text, they are able to reason better in the presence of human-induced violations, but humans find it easier to recognize and simplify violations that are unnatural.}

\eat{
\begin{figure}
    \centering
    \includegraphics[width=0.85\linewidth]{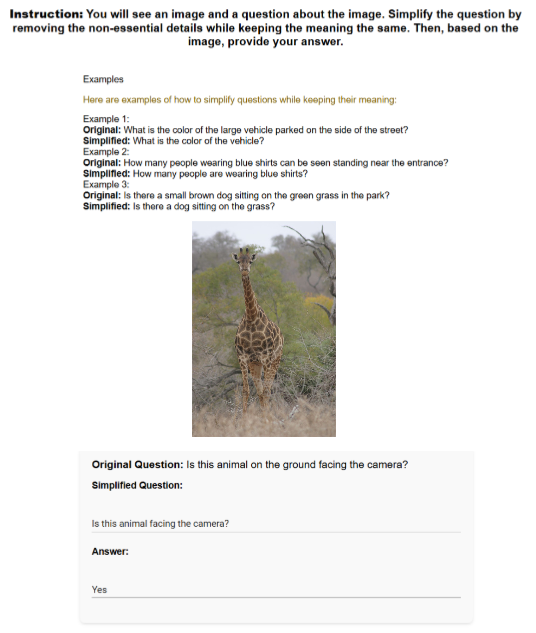}
    \caption{AMT design to measure time-on-task for humans to answer a modified question and to reflect on the answer by simplifying the question.}
    \label{fig:amt-time}
\end{figure}
}


\section{Conclusion}

In this work, we focused on evaluating the reasoning of VLMs in Visual Question Answering when they encounter violations to Grice's maxims meaning that questions could be ambiguous, overly detailed, and contain false/irrelevant information. To do this, we used VLMs themselves to add modifiers to questions and verified the modification using humans. We showed that in the presence of modifiers, VLMs are generally less accurate. Further, compared to pragmatic reasoning of humans, VLMs act more logically and when compared to VLM modifications which are unnatural, VLMs seem to perform better in human-induced modifications. In future, we plan to extend our work by considering \textit{uncertainty} of VLM generation in our analysis.
\section{Limitations}

Following are the limitations associated with this work.
\begin{enumerate}
    \item The experiments in this work assume that the original human-written questions generally follow Grice's maxims of conversation. However, it may be the case that since humans are asking a question (as opposed natural conversation when talking to fellow humans), some of these maxims are violated even in human-generated questions.
    \item Since the internal details of how VLMs handle prompts are not clearly known, there could be some bias associated with self-modified questions. That is, if a VLM tries to answer its own modified question since it would have access to the previous prompts (instructing it to add modifiers), it may be able use it in the response to modified questions. Even though we provided the modification as a separate prompt sessions, there could be some bias in the results of self-modified questions if the prompts are not completely independent.
    \item Since open-ended questions do not have a unique ground truth answer, the evaluation we used may have a human-induced bias compared to those which have a unique ground truth answer.
    \item There is inherent uncertainty in VLM responses, i.e., their responses are not deterministic in nature which could bias some of the results.In this work, we have not handled this uncertainty.
\end{enumerate}

\bibliography{custom}

@inproceedings{goyal2017making,
  title={Making the v in vqa matter: Elevating the role of image understanding in visual question answering},
  author={Goyal, Yash and Khot, Tejas and Summers-Stay, Douglas and Batra, Dhruv and Parikh, Devi},
  booktitle={Proceedings of the IEEE conference on computer vision and pattern recognition},
  pages={6904--6913},
  year={2017}
}

@incollection{grice1975logic,
  title={Logic and conversation},
  author={Grice, Herbert P},
  booktitle={Speech acts},
  pages={41--58},
  year={1975},
  publisher={Brill}
}

@article{team2024gemini,
  title={Gemini 1.5: Unlocking multimodal understanding across millions of tokens of context},
  author={Team, Gemini and Georgiev, Petko and Lei, Ving Ian and Burnell, Ryan and Bai, Libin and Gulati, Anmol and Tanzer, Garrett and Vincent, Damien and Pan, Zhufeng and Wang, Shibo and others},
  journal={arXiv preprint arXiv:2403.05530},
  year={2024}
}

@article{liu2023visual,
  title={Visual instruction tuning},
  author={Liu, Haotian and Li, Chunyuan and Wu, Qingyang and Lee, Yong Jae},
  journal={Advances in neural information processing systems},
  volume={36},
  pages={34892--34916},
  year={2023}
}

@article{hurst2024gpt,
  title={Gpt-4o system card},
  author={Hurst, Aaron and Lerer, Adam and Goucher, Adam P and Perelman, Adam and Ramesh, Aditya and Clark, Aidan and Ostrow, AJ and Welihinda, Akila and Hayes, Alan and Radford, Alec and others},
  journal={arXiv preprint arXiv:2410.21276},
  year={2024}
}

@inproceedings{yang2019auto,
  title={Auto-encoding scene graphs for image captioning},
  author={Yang, Xu and Tang, Kaihua and Zhang, Hanwang and Cai, Jianfei},
  booktitle={Proceedings of the IEEE/CVF conference on computer vision and pattern recognition},
  pages={10685--10694},
  year={2019}
}

@inproceedings{cornia2020meshed,
  title={Meshed-memory transformer for image captioning},
  author={Cornia, Marcella and Stefanini, Matteo and Baraldi, Lorenzo and Cucchiara, Rita},
  booktitle={Proceedings of the IEEE/CVF conference on computer vision and pattern recognition},
  pages={10578--10587},
  year={2020}
}

@inproceedings{wang2022end,
  title={End-to-end transformer based model for image captioning},
  author={Wang, Yiyu and Xu, Jungang and Sun, Yingfei},
  booktitle={Proceedings of the AAAI conference on artificial intelligence},
  volume={36},
  number={3},
  pages={2585--2594},
  year={2022}
}

@inproceedings{jacquet2018gricean,
  title={The gricean maxims of quantity and of relation in the turing test},
  author={Jacquet, Baptiste and Baratgin, Jean and Jamet, Frank},
  booktitle={2018 11th international conference on human system interaction (hsi)},
  pages={332--338},
  year={2018},
  organization={IEEE}
}

@inproceedings{antol2015vqa,
  title={Vqa: Visual question answering},
  author={Antol, Stanislaw and Agrawal, Aishwarya and Lu, Jiasen and Mitchell, Margaret and Batra, Dhruv and Zitnick, C Lawrence and Parikh, Devi},
  booktitle={Proceedings of the IEEE international conference on computer vision},
  pages={2425--2433},
  year={2015}
}

@inproceedings{selvaraju2020squinting,
  title={Squinting at vqa models: Introspecting vqa models with sub-questions},
  author={Selvaraju, Ramprasaath R and Tendulkar, Purva and Parikh, Devi and Horvitz, Eric and Ribeiro, Marco Tulio and Nushi, Besmira and Kamar, Ece},
  booktitle={Proceedings of the IEEE/CVF Conference on Computer Vision and Pattern Recognition},
  pages={10003--10011},
  year={2020}
}

@article{tan2019lxmert,
  title={Lxmert: Learning cross-modality encoder representations from transformers},
  author={Tan, Hao and Bansal, Mohit},
  journal={arXiv preprint arXiv:1908.07490},
  year={2019}
}

@article{davies2000grice,
  title={Grice’s cooperative principle: Getting the meaning across},
  author={Davies, Bethan},
  journal={Leeds Working Papers in Linguistics and Phonetics},
  volume={8},
  number={1},
  pages={26},
  year={2000},
  publisher={Citeseer}
}

@inproceedings{britton2022question,
  title={Question Modifiers in Visual Question Answering},
  author={Britton, William and Sarkhel, Somdeb and Venugopal, Deepak},
  booktitle={Language Resources and Evaluation Conference},
  year={2022}
}

@inproceedings{manas2024improving,
  title={Improving automatic vqa evaluation using large language models},
  author={Ma{\~n}as, Oscar and Krojer, Benno and Agrawal, Aishwarya},
  booktitle={Proceedings of the AAAI Conference on Artificial Intelligence},
  volume={38},
  number={5},
  pages={4171--4179},
  year={2024}
}

@article{zhou2023lima,
  title={Lima: Less is more for alignment},
  author={Zhou, Chunting and Liu, Pengfei and Xu, Puxin and Iyer, Srinivasan and Sun, Jiao and Mao, Yuning and Ma, Xuezhe and Efrat, Avia and Yu, Ping and Yu, Lili and others},
  journal={Advances in Neural Information Processing Systems},
  volume={36},
  pages={55006--55021},
  year={2023}
}

@article{trajman2008mcnemar,
  title={McNemar $\chi$2 test revisited: comparing sensitivity and specificity of diagnostic examinations},
  author={Trajman, A and Luiz, RR},
  journal={Scandinavian journal of clinical and laboratory investigation},
  volume={68},
  number={1},
  pages={77--80},
  year={2008},
  publisher={Taylor \& Francis}
}

@article{campbell2024understanding,
  title={Understanding the limits of vision language models through the lens of the binding problem},
  author={Campbell, Declan and Rane, Sunayana and Giallanza, Tyler and De Sabbata, Camillo Nicol{\`o} and Ghods, Kia and Joshi, Amogh and Ku, Alexander and Frankland, Steven and Griffiths, Tom and Cohen, Jonathan D and others},
  journal={Advances in Neural Information Processing Systems},
  volume={37},
  pages={113436--113460},
  year={2024}
}

@inproceedings{wang2024haloquest,
  title={Haloquest: A visual hallucination dataset for advancing multimodal reasoning},
  author={Wang, Zhecan and Bingham, Garrett and Yu, Adams Wei and Le, Quoc V and Luong, Thang and Ghiasi, Golnaz},
  booktitle={European Conference on Computer Vision},
  pages={288--304},
  year={2024},
  organization={Springer}
}

@article{wu2024autohallusion,
  title={Autohallusion: Automatic generation of hallucination benchmarks for vision-language models},
  author={Wu, Xiyang and Guan, Tianrui and Li, Dianqi and Huang, Shuaiyi and Liu, Xiaoyu and Wang, Xijun and Xian, Ruiqi and Shrivastava, Abhinav and Huang, Furong and Boyd-Graber, Jordan Lee and others},
  journal={arXiv preprint arXiv:2406.10900},
  year={2024}
}

@inproceedings{ribeiro2018semantically,
  title={Semantically equivalent adversarial rules for debugging NLP models},
  author={Ribeiro, Marco Tulio and Singh, Sameer and Guestrin, Carlos},
  booktitle={Proceedings of the 56th Annual Meeting of the Association for Computational Linguistics (volume 1: long papers)},
  pages={856--865},
  year={2018}
}

@inproceedings{
Zhang*2020BERTScore:,
title={BERTScore: Evaluating Text Generation with BERT},
author={Tianyi Zhang* and Varsha Kishore* and Felix Wu* and Kilian Q. Weinberger and Yoav Artzi},
booktitle={International Conference on Learning Representations},
year={2020},
url={https://openreview.net/forum?id=SkeHuCVFDr}
}

@inproceedings{hessel2021clipscore,
  title={CLIPScore: A Reference-free Evaluation Metric for Image Captioning},
  author={Hessel, Jack and Holtzman, Ari and Forbes, Maxwell and Le Bras, Ronan and Choi, Yejin},
  booktitle={Proceedings of the 2021 Conference on Empirical Methods in Natural Language Processing},
  pages={7514--7528},
  year={2021}
}

\section*{Appendix A: LLM Prompts, Datasets, AMT Designs, Examples}
\subsection{Prompt for answer generation from LLM}

     \textit{\uline{Instruction}}: An image is provided to you, and a question is provided as input. Provide an answer to the question based on the image. Provide mostly with a single-word response, following the instructions below:
    
    \begin{itemize}
        \item  You will be shown an image and asked a question about it.
        \item  Analyze the image carefully to identify the requested information.
        \item  Your answer should mostly be exactly one word. No explanations, no phrases, no sentences.
        \item  In cases of numbers or numeric answers, the answer should be the word form of the numbers.
        \item  If the question is about an object visible in the image, provide the single most accurate word as your answer.
    \end{itemize}
    IMPORTANT: Your answer should mostly be one word. ABSOLUTELY No explanations, no phrases, no sentences or gaps. However, it can sometimes be 2 or 3 word responses. 
    \\

\subsection{Prompt for modified question generation using VLM}

    \textit{\uline{Instruction:}} -- You are given an IMAGE and an ORIGINAL question. Create ONE modified question by ADDING a single attribute while keeping everything else from the original exactly the same. Here are the rules:
    \begin{itemize}
        \item Keep the same question type (same first question word, same format).
        \item  Keep the same target object noun from the original question. Do not replace it with a synonym or a different object.
        \item  Use exactly ONE attribute type:
    \begin{itemize}
        \item Visual attribute: insert an adjective/adjective phrase before the target noun (e.g., "brown vehicle").
        \item Relational attribute: add a short positional phrase after the target noun (e.g., "vehicle on the left").
    \end{itemize}
    \item Never use both visual and relational attributes together.
    \item The modification must NOT change the expected answer from the original question.
    \item Do not use or hint at the original answer in the modification.
    \item Use natural, grammatical language. Unrealistic attributes are allowed.
    \item Output must follow this exact format:
    \begin{itemize}
        \item Modified Questions [LIST]: [$<$modified question$>$]
        \item Attribute [LIST]: [$<$visual OR relational$>$]
    \end{itemize}
    \item Examples:
    \begin{itemize}
        \item Original: Which vehicle is closest to you?
        \item Modified Questions [LIST]: [Which brown vehicle is closest to you?]
        \item Attribute [LIST]: [visual]
    \end{itemize}
    \end{itemize}

\subsection{Prompt for modified generation from Llava}
    \textit{\uline{Instruction}} (with/without image input) -- You are a QUESTION MODIFIER, NOT a question answerer.
    \begin{itemize}
        \item DO NOT answer the question about the image.
        \item DO NOT describe what you see in the image.
         \item DO NOT tell me what objects are present.
         \item YOUR ONLY JOB: Modify the given question by adding exactly ONE attribute.
    \end{itemize}
    ORIGINAL QUESTION: "\{\}"\\
    \\
    TASK: Create a modified version by adding ONE attribute:
    \begin{itemize}
        \item Visual attribute: Add an adjective before the main noun (e.g., "red car", "large building")
        \item Relational attribute: Add a position phrase after the main noun (e.g., "car on the left", "person in the background")
    \end{itemize}
    EXAMPLES:
    \begin{itemize}
        \item Original: "What color is the car?"
        \item Modified: "What color is the red car?"
    \end{itemize}
    \begin{itemize}
        \item Original: "How many people are there?"
        \item Modified: "How many people on the left are there?"
    \end{itemize}
    \begin{itemize}
        \item Original: "Where is the dog?"
        \item Modified: "Where is the small dog?"
    \end{itemize}
    Modify the original question. Respond ONLY with:
    Modified Question: [your modified question here]
    Attribute Type: [visual or relational]

\subsection{Prompt for generating modified questions from LLM by adding hallucinated modifier}
    \textit{\uline{Instruction:}} --  
    \begin{itemize}
        \item Add a modifier to this question. 
        \item The modifier can be visual (color, size, shape, texture, pattern, visual state) OR relational (e.g., on, under, in front of, next to, in, among, etc.).
        \item Provide only ONE question with one or more modifiers.
        \item Make sure that the output is only the obtained MODIFIED QUESTION as shown below. Output Format:
        \begin{center}
            Modified Questions [LIST]: [Question1]
        \end{center}
        Example:
        \begin{itemize}
            \item Original: Is the girl in the black top thin? 
            \item Modified Questions [LIST]: [Is the girl in the black top obese?]
        \end{itemize}
    \end{itemize}
    NOTE: The response must ONLY be a question with no explanation

\subsection{Datasets}
Table~\ref{tab:dataset1-test-yesno}, \ref{tab:dataset1-test-open} are datasets sampled from the test partition of VQA2.0 used for the experiments in Section~\ref{sec:testaccuracy}. We use these to generate modified questions from each of the 4 VLMs by providing the original question and image. We also use these to generate questions using hallucinated modifications from each of the 4 VLMs by providing only the question without the image. Table~\ref{tab:dataset2-train-yesno}, \ref{tab:dataset2-train-open} show the question types in the dataset used to compare between human and AI modifications in Section~\ref{sec:trainhumans}.

\begin{table}[h!]
  \centering
  \begin{tabular}{|l|l|}
    \hline
    {\bf Question Type} & {\#Questions} \\
    \hline
     {Is the} & {75}\\
    \hline
    {Is this} & {24}\\
    \hline
    {Are the} & {24}\\
    \hline
    {Is there a} & {18}\\
    \hline
    {Is this a} & {24}\\
    \hline
    {Is the girl} & {13}\\
    \hline
    {Does the} & {23}\\
    \hline
    {Is there} & {21}\\
    \hline
    {Is the man} & {18}\\
    \hline
    {Does} & {17}\\
    \hline
    {Are they} & {18}\\
    \hline
    {Are } & {21}\\
    \hline
    {Are there} & {23}\\
    \hline
    {Are there any} & {22}\\
    \hline
    {Do} & {28}\\
    \hline
    {Is the woman} & {18}\\
    \hline
    {Are these} & {21}\\
    \hline
    {Do the} & {15}\\
    \hline
    {Is the cat} & {1}\\
    \hline
    {Are the people} & {24}\\
    \hline
    {How many} & {42}\\
    \hline
  \end{tabular}
    \caption{Total number of Open-ended-response questions types to evaluate VLM accuracy.}
\label{tab:dataset1-test-yesno}
  \end{table}

  \begin{table}[h!]
  \centering
  \begin{tabular}{|l|l|}
    \hline
    {\bf Question Type} & {\#Questions} \\
    \hline
     {What} & {351}\\
    \hline
    {Which} & {21}\\
    \hline
    {Who} & {38} \\
    \hline
    {Why} & {41}\\
    \hline
    {Where} & {63}\\
    \hline
  \end{tabular}
  \caption{Total number of Open-ended-response questions types to evaluate VLM accuracy.}
  \label{tab:dataset1-test-open}
\end{table} 
  
\begin{table}[h!]
  \centering
  \begin{tabular}{|l|l|}
    \hline
    {\bf Question Type} & {\#Questions} \\
    \hline
     {Is the} & 10\\
    \hline
    {Is this} &  57\\
    \hline
    {Are the} &  25\\
    \hline
    {Is there a} &  54\\
    \hline
    {Is this a} &  41\\
    \hline
    {Is the girl} &  1\\
    \hline
    {Does the} &  39\\
    \hline
    {Is there} &  47\\
    \hline
    {Is the man} & 9\\
    \hline
    {Does} &  30\\
    \hline
    {Are they} & 9\\
    \hline
    {Are } &  16\\
    \hline
    {Are there} &  20\\
    \hline
    {Are there any} & 8\\
    \hline
    {Do} &  12\\
    \hline
    {Is the women} &  4\\
    \hline
    {Are these} &  15\\
    \hline
    {Do the} & 3\\
    \hline
    {Is the cat} & 1\\
    \hline
    {Is the baby} & 2\\
    \hline
    {Are the people} &3\\
    \hline
    {Is the boy}& 1\\
    \hline
    {Is the lady}& 2\\
    \hline
    {Is the old man} & 1\\
    \hline
  \end{tabular}
  \caption{Total number of binary-response and quantitative-response question types in the dataset used for comparing human and AI modifications.}
  \label{tab:dataset2-train-yesno}
\end{table}

\begin{table}[h!]
  \centering
  \begin{tabular}{|l|l|}
    \hline
    {\bf Question Type} & {\#Questions} \\
    \hline
    {Which} & {41}\\
    \hline
    {Who} & 16\\
    \hline
    {Why} & 20\\
    \hline
    {Where} & {51}\\
    \hline
  \end{tabular}
  \caption{Total number of Open-ended-response questions types in the dataset used for comparing human and AI modifications.}
  \label{tab:dataset2-train-open}
\end{table} 

\subsection{Question verification using AMT}
\begin{figure*}[h!]
    \centering
    \includegraphics[width=1\linewidth]{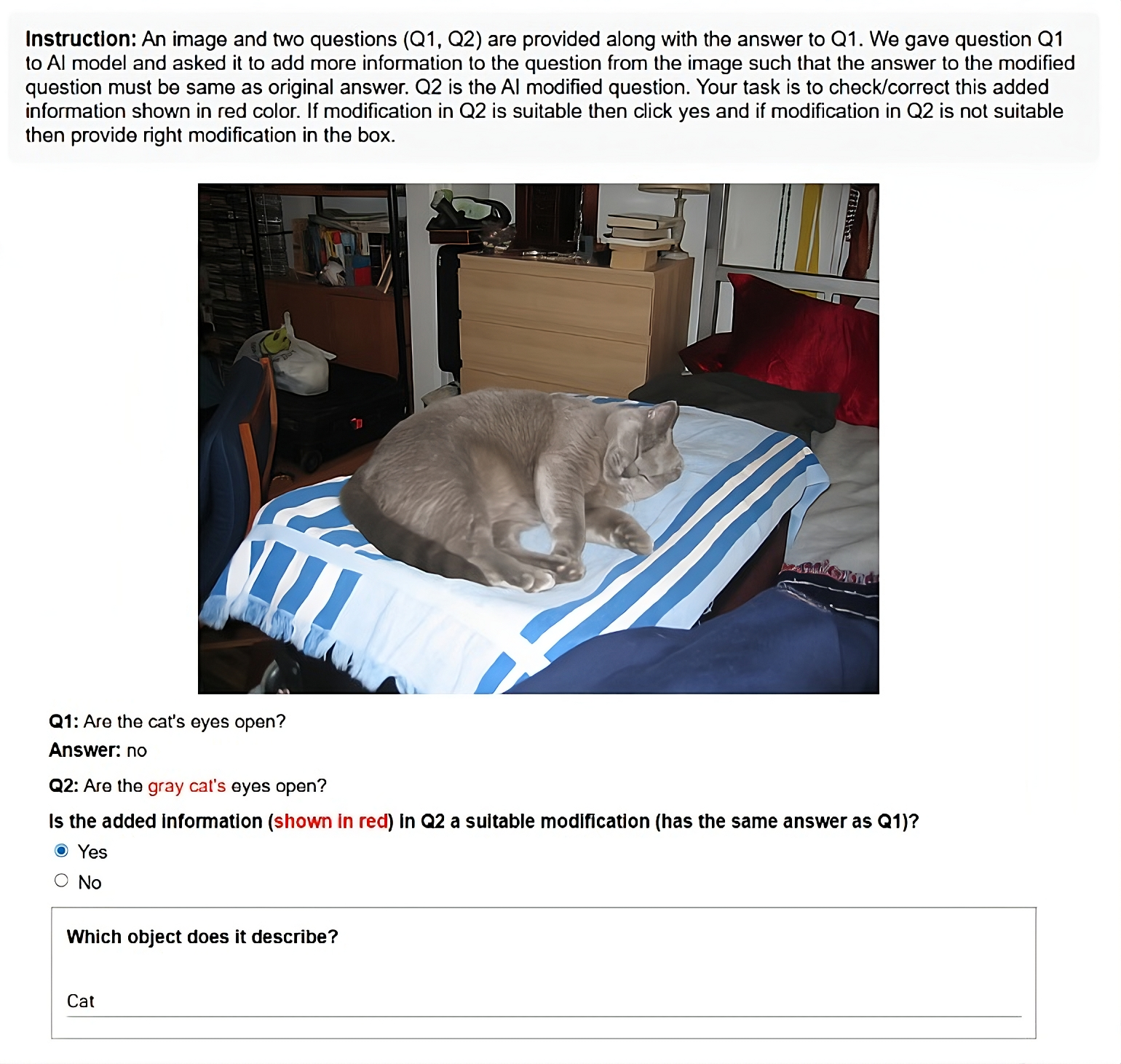}
    \caption{AMT Task for verifying if humans can pragmatically infer VLM-modified questions. If they choose that the modification is correct, they need to demonstrate their understanding of the question by answering a sub-question based on the salient object in the question.}
    \label{fig:amt-verify1}
\end{figure*}

\begin{figure*}[h!]
    \centering
    \includegraphics[width=1\linewidth]{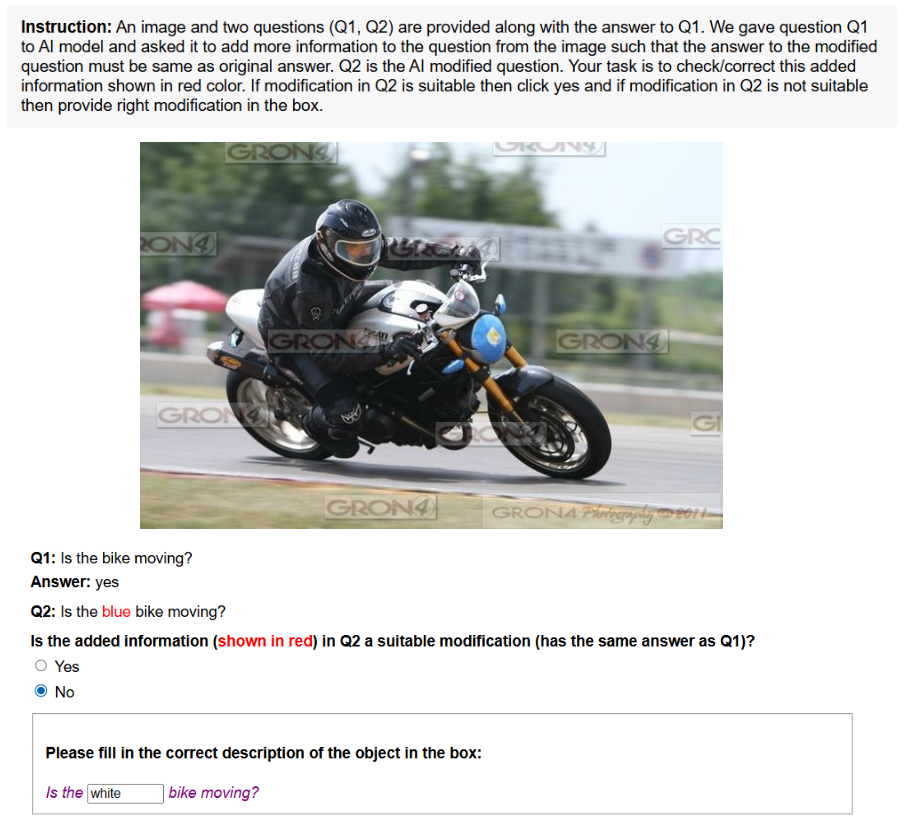}
    \caption{AMT Task for verifying if humans can pragmatically infer VLM-modified questions. If they choose that the modification is incorrect, then they need to correct the modification using a textbox as shown in the example.}
    \label{fig:amt-verify2}
\end{figure*}

\begin{figure*}[h!]
    \centering
    \includegraphics[width=1\linewidth]{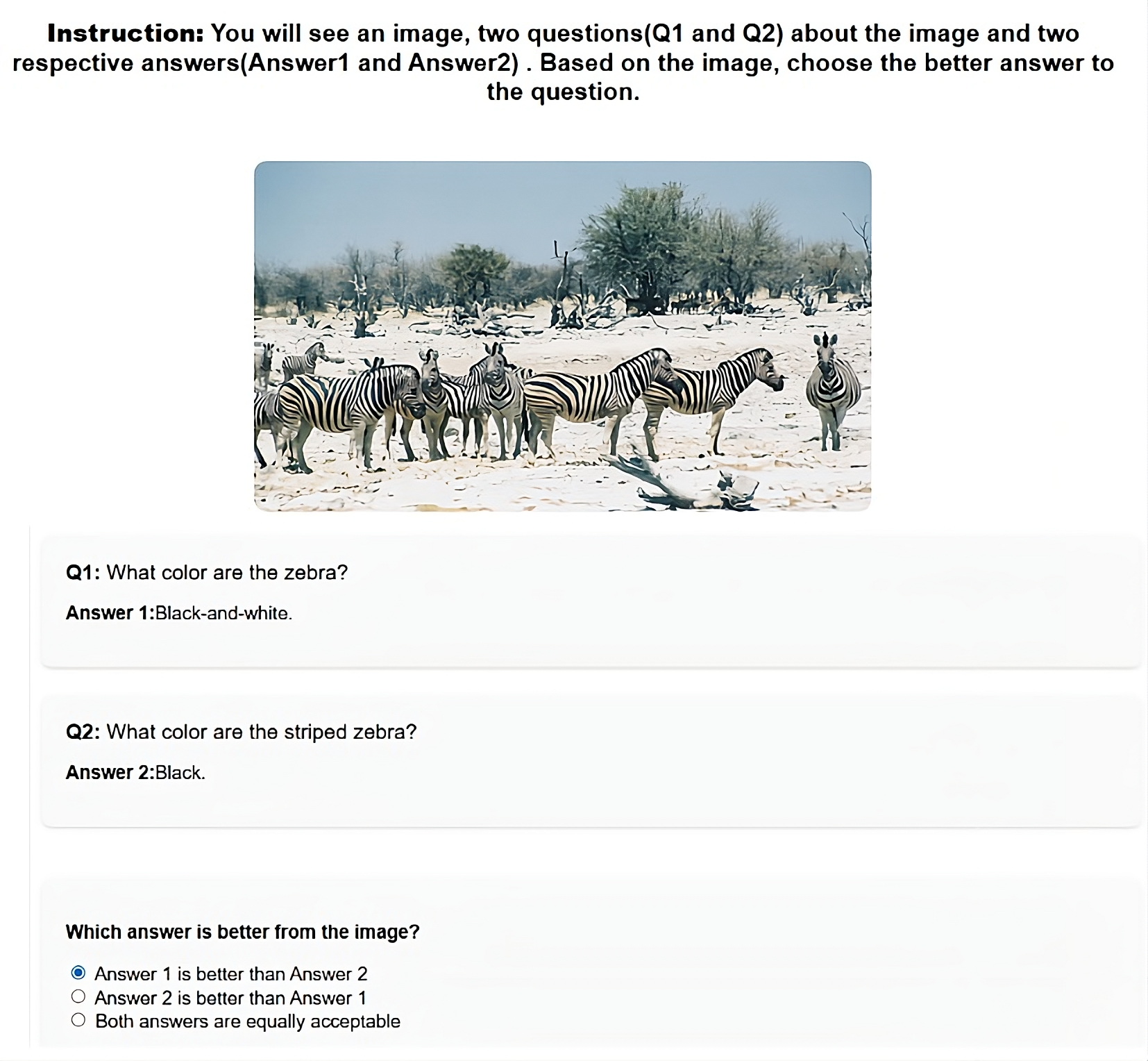}
    \caption{AMT Task for evaluating open-ended VLM responses by comparing answers to original and modified questions.}
    \label{fig:amt-compare-answers}
\end{figure*}

\begin{figure*}[h!]
    \centering
    \includegraphics[width=1\linewidth]{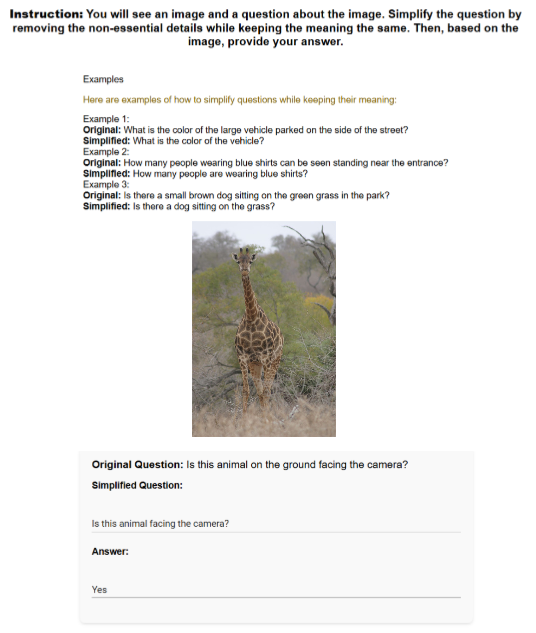}
    \caption{AMT design to measure time-on-task for humans to answer a modified question and to reflect on the answer by simplifying the question.}
    \label{fig:amt-cognitive-load}
\end{figure*}

Fig.~\ref{fig:amt-verify1}, Fig.~\ref{fig:amt-verify2} show the AMT task design for users to verify if the modification is correct or not, and then correct the modification if incorrect. Fig.~\ref{fig:amt-compare-answers} shows the AMT task to compare open-ended answers from the original and modified questions. Fig.~\ref{fig:amt-cognitive-load} shows the AMT task design for where we compute the time-on-task to measure cognitive load in processing violations within the modified questions.

\subsection{Examples}
\begin{figure*}[h!]
    \centering
    \includegraphics[width=1\linewidth]{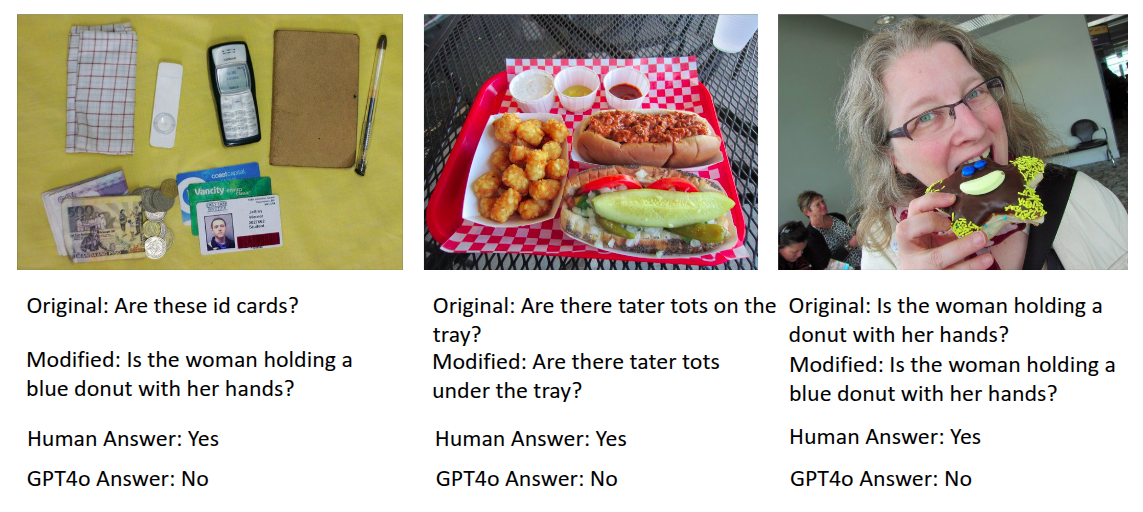}
    \caption{Examples showing human and VLM responses on hallucinated modifications}
    \label{fig:random-modifier}
\end{figure*}

\begin{figure*}[h!]
    \centering
    \includegraphics[width=1\linewidth]{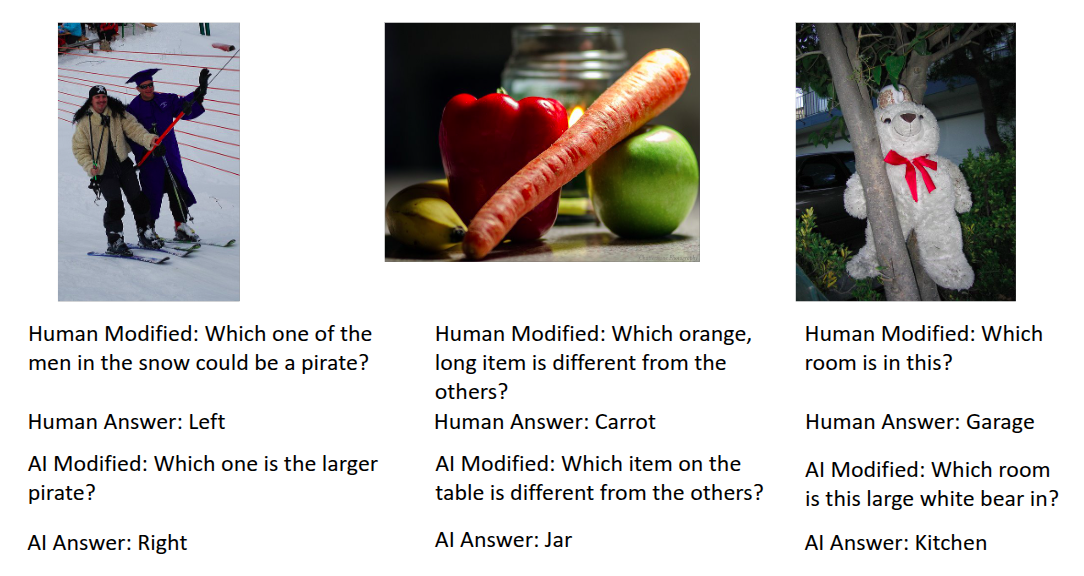}
    \caption{Examples showing human and VLM responses on human and VLM modifications}
    \label{fig:human-modifier}
\end{figure*}

Fig.~\ref{fig:random-modifier}, shows examples that compare how humans and VLMs respond to hallucinated modifiers. Fig.~\ref{fig:human-modifier} shows examples that compare how humans and VLMs respond to human and AI modified questions.

\end{document}